\documentclass[10pt,twocolumn,letterpaper]{article}

\usepackage{cvpr}              % To produce the CAMERA-READY version
\usepackage[accsupp]{axessibility}
\usepackage{multirow}
\usepackage{makecell}
\usepackage{pgfplots} %引用包
\usepackage{overpic} %图片重叠包
\usepackage{colortbl} % 引入 colortbl 宏包
\usepackage{amssymb} % 对钩
\usepackage{indentfirst} % section后首行缩进
\definecolor{cvprblue}{rgb}{0.21,0.49,0.74}
\usepackage[pagebackref,breaklinks,colorlinks,citecolor=cvprblue]{hyperref}

\def\paperID{1221} % *** Enter the Paper ID here
\def\confName{CVPR}
\def\confYear{2024}

\title{ViT-CoMer: Vision Transformer with Convolutional Multi-scale Feature Interaction for Dense Predictions}

\author
{ 
Chunlong Xia\thanks{Equal Contribution.}
\quad
Xinliang Wang$^*$
\quad
Feng Lv$^*$
\quad
Xin Hao$^*$
\quad
Yifeng Shi\thanks{Corresponding Author.}
\\
Baidu Inc.
\\
\tt\small\{xiachunlong, wangxinliang02, lvfeng02, haoxin04, shiyifeng\}@baidu.com
}

\begin{document}
\maketitle
\begin{abstract}
Although Vision Transformer (ViT) has achieved significant success in computer vision, it does not perform well in dense prediction tasks due to the lack of inner-patch information interaction and the limited diversity of feature scale. Most existing studies are devoted to designing vision-specific transformers to solve the above problems, which introduce additional pre-training costs. Therefore, we present a plain, pre-training-free, and feature-enhanced \textbf{ViT} backbone with ~\textbf{Co}nvolutional \textbf{M}ulti-scale feature int\textbf{er}action, named \textbf{ViT-CoMer}, which facilitates bidirectional interaction between CNN and transformer. Compared to the state-of-the-art, ViT-CoMer has the following advantages: (1) We inject spatial pyramid multi-receptive field convolutional features into the ViT architecture, which effectively alleviates the problems of limited local information interaction and single-feature representation in ViT. (2) We propose a simple and efficient CNN-Transformer bidirectional fusion interaction module that performs multi-scale fusion across hierarchical features, which is beneficial for handling dense prediction tasks. (3) We evaluate the performance of ViT-CoMer across various dense prediction tasks, different frameworks, and multiple advanced pre-training.~Notably, our ViT-CoMer-L achieves \textbf{64.3\% AP} on COCO val2017 \textbf{without extra training data}, and \textbf{62.1\% mIoU} on ADE20K val, both of which are comparable to state-of-the-art methods. We hope ViT-CoMer can serve as a new backbone for dense prediction tasks to facilitate future research. The code will be released at
\url{https://github.com/Traffic-X/ViT-CoMer}.
\end{abstract}
\begin{figure}[t]
\centering
\begin{overpic}[width=0.46\textwidth]{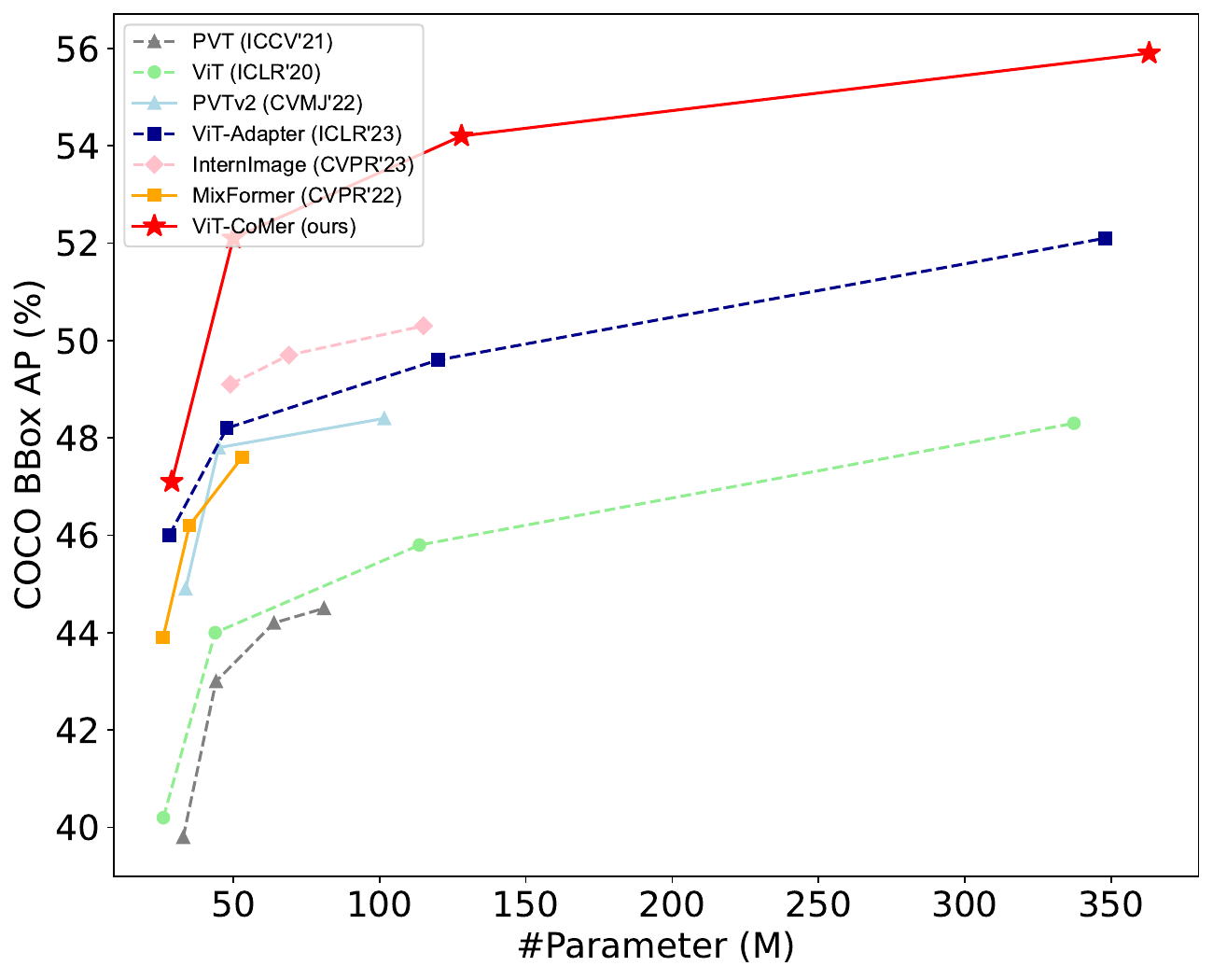}   \put(60,33){
        \tiny
        \setlength{\tabcolsep}{3pt} % 调整列间距
        \renewcommand{\arraystretch}{0.7}
        \begin{tabular}{lcc}
        \hline
        Method        & \#Param & $ \mathrm{ AP^{b} }$   \\
        \hline
        PVTv2-B1         & 34M   & 44.9 \\
        MixFormer-B3  & 35M     & 46.2 \\
        ViT-Adapter-T & 28M   & 46.0 \\
        ViT-CoMer-T   & 29M     & 47.1 \\
        \hline
        ViT-S         & 44M   & 44.0 \\
        InternImage-T & 49M     & 49.1 \\
        ViT-Adapter-S & 48M   & 48.2 \\
        ViT-CoMer-S   & 50M   & 48.8     \\
        ViT-CoMer-S\textsuperscript{$\dagger$}   & 50M   & 52.1     \\
        \hline
        ViT-B         & 114M  & 45.8 \\
        InternImage-B & 115M    & 50.3 \\
        ViT-Adapter-B & 120M  & 49.6 \\
        ViT-Adapter-B\textsuperscript{$\dagger$}   &  120M       &   51.2   \\
        ViT-CoMer-B   &  129M       &   50.2   \\
        ViT-CoMer-B\textsuperscript{$\dagger$}   &  129M       &   54.2   \\
        \hline
        ViT-L\textsuperscript{$\dagger$}         & 338M  & 48.3 \\
        ViT-Adapter-L\textsuperscript{$\dagger$} & 348M  & 52.1 \\
        ViT-CoMer-L\textsuperscript{$\dagger$}   & 363M  & 55.9 \\
        \hline
        \end{tabular}
} 
\end{overpic}
\caption{\textbf{Object detection performance on COCO val2017 using Mask R-CNN.} Our ViT-CoMer, with advanced pre-trained weights of ViT, outperforms other methods. ``\textsuperscript{$\dagger$}'' denotes the utilization of advanced pre-trained weights, otherwise ImageNet-1K.}
\vspace{-3mm}
\label{figure1}
\end{figure}
\begin{figure*}[t]
  \centering
  % \fbox{\rule{0pt}{0.5in} \rule{0.9\linewidth}{0pt}}
  % \includegraphics[width=0.8\linewidth]{egfigure.eps}
  \includegraphics[scale=0.7]{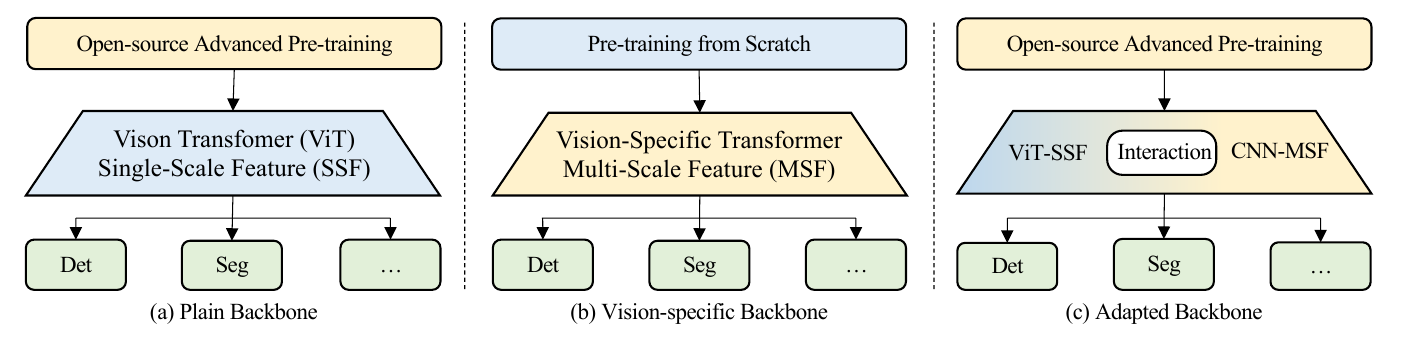}
   \caption{\textbf{Different backbone paradigms for dense predictions.} (a) Plain backbone paradigm can leverage open-source advanced pre-trained weights (e.g., BEiT series~\cite{bao2021beit, peng2022beit, beit3}, DINOv2~\cite{oquab2023dinov2}). However, its drawback lies in the limited scale diversity of feature representation, which is insufficient to meet the requirements of dense predictions. (b) Vision-specific backbone paradigm designs a multi-scale feature framework that effectively addresses dense predictions. However, each structural modification requires retraining the pre-trained weights from scratch on large-scale image datasets. (c) Adapted backbone paradigm integrates the advantages of both CNN and transformer. It can directly load advanced pre-training and achieve fusion interaction between multi-scale convolutional features and transformer features, which is beneficial for dense predictions.}
   \label{figure2}
   \vspace{-3mm}
\end{figure*}
\vspace{-3mm}
\section{Introduction}
% 近年来，随着大规模数据集的发布和深度学习的发展，在对象检测、实例分割和语义分割等密集预测任务方面取得了重大进展。这一进展导致了许多经典的卷积神经网络（CNNs）的出现，包括YOLO系列、RCNN系列、HRNet、DETR、DINO、FCN和Deeplab系列等。这些模型利用了卷积神经网络的局部连续性和多尺度能力，使其能够有效地应用于密集的预测任务。
In recent years, owing to the release of large-scale datasets~\cite{2023v2x-seq, 2022rope3d, 2022dair-v2x} and the development of deep learning~\cite{2023opentransmind}, significant progress has been made in dense prediction tasks such as object detection, instance segmentation, and semantic segmentation (e.g.,YOLO series~\cite{2017YOLO9000,redmon2016yolov1,yolov3,2020YOLOv4}, RCNN series~\cite{2017Mask-rcnn,cai2019cascade-rcnn}, DETR~\cite{carion2020detr}). This progress has led to the emergence of numerous classic convolutional neural networks (CNNs), including ResNet~\cite{he2016resnet}, ConvNeXt~\cite{liu2022convnext}, etc. These models leverage the local continuity and multi-scale capabilities of convolutional neural networks, enabling them to be effectively applied to dense prediction tasks.
% 受到transformer在NLP领域取得的效果的激发，ViT作为第一个将transformer应用在视觉分类任务的算法，同样取得了引人注意的效果。之后越来越多的研究人员将ViT-Based模型应用于密集预测任务中。目前，针对密集预测任务所设计的基于Transformer网络结构主要分为三种范式，见图1. plain backbone，vision-specific backbone 和 adapted backbone. plain backbone方法是在不改变vit的框架，优化ViT feature 使用方式，例如vitdet，见图一（a)。vision-specific backbone （比如Swin, PVT）结合了CNN与transformer的优势重新搭建网络结构，从而更好地完成密集预测任务。adapted backbone 的特点是不改变vit结构，仅通过添加额外分支引入CNN特征，同时可直接加载各种各样的开源的vit预训练权重，提升vit在密集预测任务上的表现。
Meanwhile, inspired by the success of transformers in NLP, the Vision Transformer (ViT)~\cite{dosovitskiy2020vit} garners significant attention as the pioneering approach to applying transformers to visual tasks. Currently, Transformer-based network architecture designed for dense prediction tasks is mainly divided into three paradigms: plain backbone, vision-specific backbone, and adapted backbone, as shown in Figure~\ref{figure2}. The plain backbone optimizes the use of ViT features without changing the framework of ViT, such as ViTDet~\cite{li2022vitdet}, as shown in Figure~\ref{figure2}(a). The vision-specific backbone (e.g., Swin~\cite{liu2021swin}, CMT~\cite{guo2022cmt}, MPViT~\cite{lee2022mpvit}, PVT series~\cite{wang2021pvt, wang2022pvtv2}), combines the advantages of CNN and Transformer to redesign the network structure, which helps them achieve better performance in dense prediction tasks, as shown in Figure~\ref{figure2}(b). The adapted backbone, shown in Figure~\ref{figure2}(c), is based on the plain ViT, which only introduces CNN features by adding additional branches and can directly load various open-source and powerful ViT pre-trained weights to improve ViT performance on dense prediction tasks.
% in this work，我们提出了一种朴素的，无需预训练的，特征增强的ViT框架，名字叫做ViT-Comer, 它只增加了少数参数量，并且可以直接加载各类预训练权重。 具体来说，我们设计了2个核心模块:多尺度多感受野特征金字塔模块和CNN-transformer双向融合交互模块。前者能够为ViT提供了更加丰富的空间信息；后者则能够融合多尺度卷积特征和tranformer特征，从而具备更强大的表征能力。ViT-coMer在权重初始化过程中，ViT直接采用开源的预训练权重，剩余模块采用随机初始化设置。结果见图一，从图中我们可以看出，当使用更好的ViT的预训练权重，我们的模型有更好的表现。当预训练权重保持一致时，ViT-CoMer也可以获得比现有方案更好的结果。我们的重要贡献如下所示：
In this work, we present a plain, pre-training-free, and feature-enhanced ViT backbone named ViT-CoMer, which can directly load various open-source and advanced pre-trained weights. Specifically, we design two core modules: the \textbf{M}ulti-\textbf{R}eceptive Field \textbf{F}eature \textbf{P}yramid module (MRFP) and the \textbf{C}NN-\textbf{T}ransformer Bidirectional Funsion \textbf{I}nteraction module (CTI). MRFP can supplement ViT with more abundant multi-scale spatial information; CTI can fuse multi-scale features from CNN and Transformer, facilitating the model with a more powerful feature representation ability. In the weight initialization process of ViT-CoMer, the ViT module directly uses the open-source pre-training, and the rest use random initialization. As shown in Figure~\ref{figure1}, our model performs better when using advanced pre-trained weights of ViT. Our main contributions are as follows:
\begin{itemize}
\item We propose a novel dense prediction backbone by combining the plain ViT with CNN features. It effectively leverages various open-source pre-trained ViT weights and incorporates spatial pyramid convolutional features that address the lack of interaction among local ViT features and the challenge of single-scale representation.
% 2）我们设计了多感受野特征金字塔模块和CNN-transformer双向交互模块。他们不仅可以对彼此的特征进行了丰富与增强，同时层级特征之间还进行多尺度融合，得到了更加丰富的语义信息，有利于处理密集预测任务。
\item We design a multi-receptive field feature pyramid module and a CNN-Transformer bidirectional fusion interaction module. The former can capture various spatial features, the latter performs multi-scale fusion across hierarchical features to obtain richer semantic information, which is beneficial for handling dense prediction tasks.
% 3）我们在COCO和ADE20K benchmark上评估了我们的ViT_CoMer,   在COCO val上，ViT-Comer使用相同的pre-training，在目标检测和实例分割任务中优于现有的同类型sota方案。令人惊喜的是，vit-comer-Small模型可以获得比vit-large更好的检测结果，且它的参数量不足vit-large参数量的1/6。进一步的，当使用大规模多模态预训练时，同等参数量下，vit-comer获得了和vision-specific 的sota检测模型相媲美的效果。在ADE20K val上，vit-comer同样可以获得比同类型sota方案更好的结果。
\item We evaluate our proposed ViT-CoMer on several challenging dense prediction benchmarks, including object detection, instance segmentation and semantic segmentation. The experimental results demonstrate that our method significantly enhances the capabilities of the plain ViT. Especially, when utilizing advanced open-source pre-training such as DINOv2~\cite{oquab2023dinov2}, ViT-CoMer can consistently outperform SOTA methods under fair comparison conditions. Notably, our ViT-CoMer-L, with advanced pre-training, achieves \textbf{64.3\% AP}, which is the best record on COCO val2017 \textbf{without training on extra detection data (e.g., Objects365)}. Moreover, ViT-CoMer-L attains \textbf{62.1\% mIoU} on the ADE20K val, which is comparable with SOTA methods.
\end{itemize}
\begin{figure*}[t]
  \centering
  % \fbox{\rule{0pt}{0.5in} \rule{0.9\linewidth}{0pt}}
  % \includegraphics[width=0.8\linewidth]{egfigure.eps}
  \includegraphics[width=1.0\linewidth]{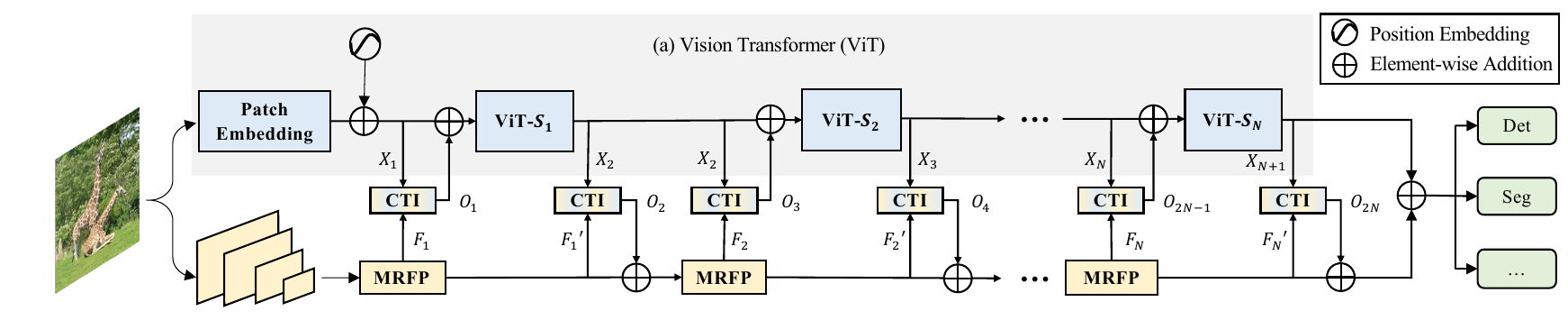}
   \caption{\textbf{The overall architecture of ViT-CoMer.} ViT-CoMer is a two-branch architecture consisting of three components: (a) a plain ViT with L layers, which is evenly divided into N stages for feature interaction. (b) a CNN branch that employs the proposed \textbf{M}ulti-\textbf{R}eceptive Field \textbf{F}eature \textbf{P}yramid (MRFP) module to provide multi-scale spatial features, and (c) a simple and efficient \textbf{C}NN-\textbf{T}ransformer Bidirectional Fusion \textbf{I}nteraction (CTI) module to integrate the features of the two branches at different stages, enhancing semantic information.}
   \label{figure3}
\end{figure*}

% \vspace{-1mm}
\section{Related Work}
% VIT首次将transformer结构引入到计算机视觉的分类任务中，并取得了令人惊喜的结果。vit是一个直筒的，单一尺度的的模型架构。 ViTDet是一个目标检测算法，该算法基于原始的ViT结构增加了简单的特征金字塔模块，预测结果相比较当前的SOTA方案还有一定的差距。可能的原因是vit特征不够丰富。密集预测模型需要具备良好的多尺度感知能力。our work 将多尺度增强卷积特征与vit特征相结合，使得模型在处理密集预测任务时能够提取出丰富的多尺度特征。
\subsection{Plain backbones} ViT~\cite{dosovitskiy2020vit} is the first work introducing the transformer to the image classification task and achieving impressive results. ViTDet~\cite{li2022vitdet} is a plain, non-hierarchical detector based on ViT by incorporating a simple feature pyramid module. However, the performance of ViTDet exhibits a gap compared to state-of-the-art methods. One potential reason is that the feature representation of ViT might not be sufficiently rich. Nonetheless, dense prediction models require a strong ability for multi-scale perception. Our work combines multi-scale enhanced convolutional features with ViT features, enabling the model to extract rich multi-scale features when dealing with dense prediction tasks.
% vision-specific backbone 主要是为了解决直筒vit特征尺度单一，计算开销大，局部特征无信息交互等问题。Swin-Transformer算法利用滑动窗口操作缓解了vit局部信息无交互的问题，同时构建了多尺度特征以适配密集目标预测问题。同时swin-transformer降低了整个框架的计算量。PVT算法构建了特征金字塔结构，重点解决了vit单一尺度特征的问题，同时对于transformer的计算方式进行了简化，有效的降低的算法的计算量。mixformer利用了cnn网络和transformer网络的优势，通过双向交互模块进行vit特征和cnn特征的相互增强，同时它也是一种多尺度特征网络结构。对包含密集预测在类的视觉任务都有一个不错的效果。iformer从高低频的角度分析了cnn和transformer的优势，设计了一种同时包含卷积和transformer的一种混合模块。并基于该模块设计了一种多尺度网络模型，iformer在分类分割检测等任务中取得了不错的效果。metaformer提出了一种通用的层级网络架构，利用pooling代替attention在分类检测分割任务中获得了较好的结果。uniformer算法也是一篇集成cnn和transformer结构优势的论文，通过在同一个block里级联cnn和attention结构获得两者特征表示能力，同时构建了特征金字塔网络框架，在目标检测中取得了不错的效果。vision-specific backbones 改变了vit结构，它们无法直接使用现成的更强大的预训练权重，如imagenet-22k，或多模态预训练。our work不破坏原始vit结构，因此可以直接使用开源的基于vit的预训练模型，使得模型快速具备更强的泛化性能与多模态能力。
\subsection{Vision-specific backbones} Vision-specific backbones are primarily designed to alleviate challenges in ViT, such as the non-hierarchical feature, and the lack of interaction among local features. Swin-Transformer~\cite{liu2021swin} employs shifted windows to alleviate the lack of interaction among local information in ViT. Simultaneously, it constructs multi-scale features to adapt the dense predictions. PVT~\cite{wang2021pvt} constructs a feature pyramid structure to address the limitations of single-scale features in ViT, which simplifies the structure of the transformer and effectively reduces the computational complexity. MixFormer~\cite{chen2022mixformer} utilizes a bidirectional feature interaction operator with convolution and self-attention to enhance feature representation. iFormer~\cite{si2022iformer} analyzes the advantages of CNN and Transformer architectures at high and low frequencies. MetaFormer~\cite{yu2022metaformer} introduces a general hierarchical network architecture that utilizes pooling instead of attention, which achieves favorable results in various vision tasks. UniFormer~\cite{li2022uniformer} cascades CNN and attention within a block, which integrates the advantages of both CNN and Transformer. Vision-specific backbones alter the ViT structure, which prevents them from directly using existing powerful pre-trained weights, e.g., BEiT series~\cite{beit3,bao2021beit, peng2022beit}. Our work preserves the original ViT, allowing it to load open-source pre-trained weights based on ViT directly. This enables our model to rapidly acquire enhanced generalization performance.
% vit-adapter提出了一种结合空间先验信息的vit框架，它利用了vit预训练权重的优势，然后通过空间先验信息增强vit特征，训练阶段full-finetune整个网络，在密集预测任务中取得了不错的效果。该方案缺少空间先验信息之间的特征交互。VPT提出了一种冻结vit预训练权重，只对adapter模块进行训练更新，在部分任务中可以获得和full-finetune方式相比较的结果，但是在语义分割任务中效果不如full-finetune的方式，该方案的优点是训练参数较full-finetune方式大幅度减少。LoRand也是一种保持vit权重不变，只训练adapter模块的一种算法，该算法训练参数极少。只占整体训练参数的1%-3%，在密集预测任务中效果没有full-finetune效果好。我们的工作对于空间层级特征本身做了特征融合增强，在训练阶段采用了full-finetune的方式优化模型，有效的提升了算法的整体效果。
\subsection{Adapted backbones} ViT-Adapter~\cite{chen2022vit-adapter} presents a ViT framework that integrates spatial prior information. It leverages the advantages of ViT's pre-trained weights. ViT-adapter needs to full-finetune during training, resulting in an impressive performance in dense prediction tasks. Meanwhile, it lacks feature interaction among spatial prior information. VPT~\cite{jia2022vpt} introduces a method that freezes the pre-trained weights of ViT and updates only the parameters of the adapter module during training. While this approach can yield results comparable to the full-finetuning method in some tasks, it doesn't perform as well as full-finetuning in semantic segmentation. LoRand~\cite{yin2023lorand} is also an algorithm that preserves the weights of ViT and trains only the adapter module. which only need to train 1\%–3\% of the overall training parameters. However, its performance is not as effective as the full-finetuning approach. Our work enhances spatial hierarchical features through feature fusion and employs the full-finetuning approach to optimize the model during training, which effectively boosts the performance of the model.
\section{The ViT-CoMer Method}
\subsection{Overall Architecture}

 The overall architecture of ViT-CoMer is illustrated in Figure~\ref{figure3}, which includes three parts: (a) Plain ViT. (b) Multi-receptive field feature pyramid module\space(MRFP). (c) CNN-Transformer bidirectional fusion interaction module\space(CTI). Firstly, for the ViT branch (see Figure~\ref{figure3}(a)), an input image with the shape of  $H\times W \times 3$ is fed into the patch embedding to obtain features with a resolution reduction of $1/16$ of the original image. Meanwhile, for the other branch, this image passes through a stack of convolutions to obtain feature pyramid $C_{3} $, $C_{4} $, and $C_{5} $ with resolutions of $1/8$, $1/16$, and $1/32$, and each of them contains D-dimensional feature maps. Secondly, both of the two branch features pass through N stage feature interactions. At each stage, the feature pyramid will first be enhanced through the MRFP module, and then bidirectionally interact with the feature of ViT through the CTI module, which can obtain multi-scale features with rich semantic information. CTI operates at the beginning and end of each stage. After N stage feature interactions, the features from two branches are added at each scale for dense prediction tasks.

\subsection{Multi-Receptive Field Feature Pyramid}\label{3.1}
\begin{figure}[t]
  \centering
  % \fbox{\rule{0pt}{0.5in} \rule{0.9\linewidth}{0pt}}
  % \includegraphics[width=0.8\linewidth]{egfigure.eps}
  % \includegraphics[width=1.0\linewidth]{AnonymousSubmission/LaTeX/figs/vit-mrfp-16.png}
  \includegraphics[scale=0.58]{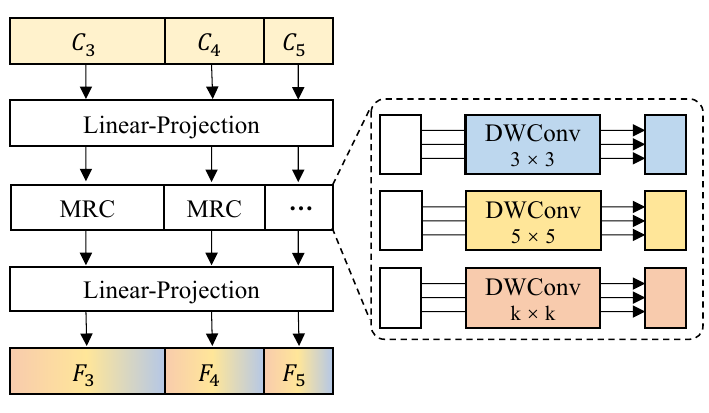}
   \caption{\textbf{Multi-Receptive Field Feature Pyramid module.} The $C_{3}$, $C_{4}$, and $C_{5}$ features are first dimensionally reduced through a linear projection layer. Subsequently, these features are divided into multiple groups along the channel dimension. Different groups employ varied kernel sizes of DWConv to enrich receptive field representation, MRC represents a multi-receptive field convolution operation. Finally, the features are restored to their original dimensions through dimensional expansion.}
   \label{figure4}
\end{figure}

The multi-receptive field feature pyramid module consists of a feature pyramid and multi-receptive field convolutional layers. The feature pyramid can provide rich multi-scale information, while the latter expands the receptive field through different convolution kernels, enhancing the long-range modeling ability of CNN features. The module is shown in Figure~\ref{figure4}. MRFP is composed of two linear projection layers and a set of depth-wise separable convolutions with multi-receptive fields. Specifically, the input of the module is a set of multi-scale features $\left\{C_{3}, C_{4}, C_{5}\right\}$, we flatten and concatenate these feature maps into feature tokens  $C \in \mathbb{R}^{(\frac{HW}{8^2} + \frac{HW}{16^2} + \frac{HW}{32^2} )\times D}$, which first passes through a linear projection layer to obtain dimensionally reduced features, and then the features are divided into $M$ groups on the channel dimension. Different groups of features correspond to convolutional layers with different receptive fields $(e.g., k = 3\times3, 5\times5)$. Finally, the processed features are concatenated and dimensionally increased through the linear projection layer. The process can be represented as:
\begin{equation}
\mathit{F} = \mathit{FC}(\mathit{DWConv(FC({C}))}),
\end{equation}
where $FC(\cdot)$ is linear projection, $DWConv(\cdot)$ is a set of depth-wise convolutions with different kernel sizes.

% \vspace{2mm}
\subsection{CNN-Transformer Bidirectional Fusion Interaction}
\begin{figure}[t]
  \centering
  % \fbox{\rule{0pt}{0.5in} \rule{0.9\linewidth}{0pt}}
  % \includegraphics[width=0.8\linewidth]{egfigure.eps}
  \includegraphics[scale=0.6]{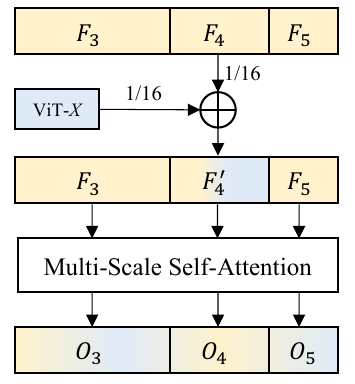}
   \caption{\textbf{CNN-Transformer Bidirectional Fusion Interaction module.} %It adopts the fusion method of direct addition, which is simple and effective. Then it uses multi-scale self-attention to unify the two modality features of CNN and Transformer, and finally completes the interaction and obtains updated features.%
   \{$F_{3}$, $F_{4}$, $F_{5}$\} are multi-scale CNN features obtained through the MRFP module. We add $F_{4}$ and $X$ from the ViT branch and use a multi-scale self-attention module to unify the two modal features, ultimately achieving information interaction and obtaining updated features.}
   \label{figure5}
\end{figure}

% \vspace{2mm}
% 我们提出了一种CNN-Transformer Bidirectional Fusion Interaction模块，如图5所示。在不改变vit结构的前提下，它通过跨架构融合的方式，引入CNN多尺度特征信息。然后通过双向交互的方式缓解了ViT特征尺度单一以及缺乏局部信息交互的问题，同时提升了CNN长距离建模能力，使其具备更强大的语义表达能力。
We propose a cross-architecture feature fusion method named CTI, as shown in Figure~\ref{figure5}. It introduces CNN's multi-scale features without altering the ViT structure. Simultaneously, through bidirectional interaction, we alleviate the problems of the lack of inner-patch information interaction and the non-hierarchical feature in ViT, while further enhancing the CNN's long-range modeling ability and semantic representation.
%  对于ViT输入特征x「HWC」和经过MRFP模块后的多尺度特征C[c3 c4 c5]，我们采用了一种简单有效的融合方式，将具有同尺度的特征x和c4直接相加，得到集合$C',表示为,它聚合了来自不同架构的多种尺度特征。但由于架构的差异，它们在模态的表达上有不同的偏向性（如高低频语义、全局局部信息）。 因此，我们采用self-attention来实现CNN与Transformer表征统一。过程可以表示为：
% 同时，由于C′包含了下采样为1/8,1/16,1/32的多种尺度特征，在self-attention过程中可以实现不同尺度patch之间的交互，使得模型能够更好地捕捉到图像中的多尺度信息，这与传统的Transformer结构完全不同，后者仅在单尺度上使用self-attention。最后，通过有效地融合多尺度CNN与Transformer特征，能够使得模型具备更强大的建模能力。
% \subsubsection{Cross-Architecture Fusion.}  
In order to fuse the ViT feature $X \in \mathbb{R}^{\frac{H}{16} \times \frac{W}{16} \times D}$ and the multi-scale feature $\left\{F_{3}, F_{4}, F_{5}\right\}$ obtained through the MRFP module, it can be represented as $F \in \mathbb{R}^{(\frac{HW}{8^2} + \frac{HW}{16^2} + \frac{HW}{32^2}) \times D}$. We directly add features $X$ and $F_{4}$, yielding the set $F'$, expressed as $F'= \left\{F_{3}, F'_{4}, F_{5}\right\}$, which aggregates multi-scale features from different architectures. However, due to architectural differences, they exhibit bias in modality representation (e.g., high-low frequency semantics, and global-local information). To address this, we employ self-attention to unify CNN and Transformer features, reinforcing the representation invariance against modality discrepancy. The process can be described as: 
\begin{equation}
\mathit{O} = \mathit{FFN}(\mathit{Attention(norm({F'}))}),
\end{equation}
where the $norm(\cdot)$ is LayerNorm~\cite{ba2016layer}, $Attention(\cdot)$ is multi-scale deformable attention~\cite{zhu2020deformable} and $\mathit{FFN}(\cdot)$ is feed-forward network. Finally, we align the feature map sizes of $O_{3}$ and $O_{5}$ to $O_{4}$ through bilinear interpolation and add $X$ as the input of the next ViT layer. Moreover, since $F'$ contains multi-scale features with resolutions of $1/8, 1/16,$ and $1/32$, self-attention can facilitate interaction among multi-scale features, and enable the model to better capture multi-scale information in images. This diverges from the traditional transformer architecture, which employs self-attention solely on a single-scale feature. By effectively fusing multi-scale CNN and Transformer features, the model gains enhanced modeling capability.
% \begin{small}
\begin{table*}[ht!]
\centering
\small
\setlength{\tabcolsep}{3.5pt} % 调整列间距
\begin{tabular}{l | c | cccccc | cccccc}
%\begin{tabular}{{p{2.2cm}|p{1.cm}||P{0.5cm}P{0.5cm}P{0.5cm}P{0.5cm}P{0.5cm}P{0.5cm}||P{0.5cm}P{0.5cm}P{0.5cm}P{0.5cm}P{0.5cm}P{0.5cm}}}
\hline
\multirow{2}{*}{Method} &
  \multirow{2}{*}{\#Param} &
  \multicolumn{6}{c|}{Mask R-CNN 1× schedule} &
  \multicolumn{6}{c}{Mask R-CNN 3× schedule} \\
 &
   &
  $ \mathrm{ AP^{b} }$ &
  $ \mathrm{ AP^{b}_{50} }$ &
  $ \mathrm{ AP^{b}_{75} }$ &
  $ \mathrm{ AP^{m} }$ &
  $ \mathrm{ AP^{m}_{50} }$ &
  $ \mathrm{ AP^{b}_{75} }$ &
  $ \mathrm{ AP^{b} }$ &
  $ \mathrm{ AP^{b}_{50} }$ &
  $ \mathrm{ AP^{b}_{75} }$ &
  $ \mathrm{ AP^{m} }$ &
  $ \mathrm{ AP^{m}_{50} }$ &
  $ \mathrm{ AP^{b}_{75} }$ \\ \hline
PVT-T~\cite{wang2021pvt}      & 33M  & 36.7 & 59.2 & 39.3 & 35.1 & 56.7 & 37.3 & 39.8 & 62.2 & 43.0 & 37.4 & 59.3 & 39.9 \\
%PVTv2-B1~\cite{wang2022pvtv2}      & 33.7M  & 41.8 & 64.3 & 45.9 & 38.8 & 61.2 & 41.6 & 44.9 & 67.3 & 49.4 & 40.8 & 64.0 & 43.8 \\
ViT-Adapter-T~\cite{chen2022vit-adapter} & 28M  & 41.1 & 62.5 & 44.3 & 37.5 & 59.7 & 39.9 & 46.0 & 67.6 & 50.4 & 41.0 & 64.4 & 44.1 \\
ViT-T~\cite{li2021benchmarking}         & 26M  & 35.5 & 58.1 & 37.8 & 33.5 & 54.9 & 35.1 & 40.2 & 62.9 & 43.5 & 37.0 & 59.6 & 39.0 \\
ViTDet-T~\cite{li2022vitdet}      & 27M  & 35.7 & 57.7 & 38.4 & 33.5 & 54.7 & 35.2 & 40.4 & 63.3 & 43.9 & 37.1 & 60.1 & 39.3 \\
PVT-S~\cite{wang2021pvt}   & 44M  & 40.4 & \textbf{62.9} & 43.8 & 37.8 & \textbf{60.1} & 40.3 & 43.0 & 65.3 & 46.9 & 39.9 & 62.5 & 42.8 \\
\rowcolor[HTML]{ECF4FF}
ViT-CoMer-T (ours)  & 29M  & \textbf{42.1} & 62.7 & \textbf{45.3} & \textbf{38.0} & \textbf{60.1} & \textbf{40.5} &  \textbf{47.1} & \textbf{67.8} &   \textbf{51.5} & \textbf{41.5} & \textbf{64.8}  & \textbf{44.3} \\ \hline
ConvNeXt-T~\cite{liu2022convnext}    & 48M  & 44.2 & 66.6 & 48.3 & 40.1 & 63.3 & 42.8 & 46.2 & 67.9 & 50.8 & 41.7 & 65.0 & 44.9 \\
Focal-T~\cite{yang2021focal}       & 49M  & 44.8 & 67.7 & 49.2 & 41.0 & 64.7 & 44.2 & 47.2 & 69.4 & 51.9 & 42.7 & 66.5 & 45.9 \\
SPANet-S~\cite{yun2023spanet} & 48M  & 44.7 & 65.7 & 48.8 & 40.6 & 62.9 & 43.8 & - & - & - & - & - & - \\
MixFormer-B4~\cite{chen2022mixformer}  & 53M  & 45.1  & 67.1 & 49.2 & 41.2 & 64.3 & 44.1 & 47.6   & 69.5   & 52.2   & 43.0   &  66.7     & 46.4   \\
ViT-Adapter-S~\cite{chen2022vit-adapter} & 48M  & 44.7 & 65.8 & 48.3 & 39.9 & 62.5 & 42.8 & 48.2 & 69.7 & 52.5 & 42.8 & 66.4 & 45.9 \\
Twins-B~\cite{chu2021twins} & 76M  & 45.2 & 67.6 & 49.3 & 41.5 & 64.5 & 44.8 & 48.0 & 69.5 & 52.7 & 43.0 & 66.8 & 46.6 \\
Swin-S~\cite{liu2021swin} & 69M  & 44.8 & 66.6 & 48.9 & 40.9 & 63.4 & 44.2 & 48.5 & 70.2 & 53.5 & 43.3 & 67.3 & 46.6 \\
Flatten-PVT-T~\cite{han2023flatten} & 49M  & 44.2 & 67.3 & 48.5 & 40.2 & 63.8 & 43.0 & 46.5 & 68.5 & 50.8 & 42.1 & 65.4 & 45.1 \\
NAT-S~\cite{hassani2023neighborhood} & 70M  & - & - & - & - & - & - & 48.4 & 69.8 & 53.2 & 43.2 & 66.9 & 46.5 \\
ViT-S~\cite{li2021benchmarking}         & 44M  & 40.2 & 63.1 & 43.4 & 37.1 & 60.0 & 38.8 & 44.0 & 66.9 & 47.8 & 39.9 & 63.4 & 42.2 \\
ViTDet-S~\cite{li2022vitdet}      & 46M  & 40.6 & 63.3 & 43.5 & 37.1 & 60.0 & 38.8 & 44.5 & 66.9 & 48.4 & 40.1 & 63.6 & 42.5 \\
\rowcolor[HTML]{ECF4FF}
ViT-CoMer-S~(ours)  &  50M  &  45.8  & 67.0 & 49.8 & 40.5 & 63.8 & 43.3 & 48.8 & 69.4 &  53.5  & 43.0 & 66.9 & 46.3 \\
\rowcolor[HTML]{DAE8FC}
ViT-CoMer-S\textsuperscript{$\ddagger$}~(ours)  &  50M  &  \textbf{48.6}  & \textbf{70.5} & \textbf{53.1} & \textbf{42.9} & \textbf{67.0} & \textbf{45.8} & \textbf{52.1} & \textbf{73.1} &    \textbf{57.1}  & \textbf{45.8} & \textbf{70.2} & \textbf{49.4} \\ \hline
PVTv2-B5~\cite{wang2022pvtv2}     & 102M & 47.4 & 68.6 & 51.9 & 42.5 & 65.7 & 46.0 & 48.4 & 69.2 & 52.9 & 42.9 & 66.6 & 46.2 \\
Swin-B~\cite{liu2021swin}        & 107M & 46.9 & -    & -    & 42.3 & -    & -    & 48.6 & 70.0 & 53.4 & 43.3 & 67.1 & 46.7 \\
InternImage-B~\cite{wang2023internimage} & 115M & 48.8 & 70.9 & 54.0 & 44.0 & 67.8 & 47.4 & 50.3 & 71.4 & 55.3 & 44.8 & 68.7 & 48.0 \\
ViT-Adapter-B~\cite{chen2022vit-adapter} & 120M & 47.0 & 68.2 & 51.4 & 41.8 & 65.1 & 44.9 & 49.6 & 70.6 & 54.0 & 43.6 & 67.7 & 46.9 \\
ViT-B~\cite{li2021benchmarking}         & 114M & 42.9 & 65.7 & 46.8 & 39.4 & 62.6 & 42.0 & 45.8 & 68.2 & 50.1 & 41.3 & 65.1 & 44.4 \\
ViTDet-B~\cite{li2022vitdet}      & 121M & 43.2 & 65.8 & 46.9 & 39.2 & 62.7 & 41.4 & 46.3 & 68.6 & 50.5 & 41.6 & 65.3 & 44.5 \\
\rowcolor[HTML]{ECF4FF}
ViT-CoMer-B~(ours)         &  129M   & 47.6 & 68.9 & 51.9 & 41.8 & 65.9 & 44.9 & 50.2 & 70.7 & 54.9 & 44.0 & 67.9 & 47.4  \\ 
\rowcolor[HTML]{DAE8FC}
ViT-CoMer-B\textsuperscript{$\ddagger$}~(ours)         &  129M   & \textbf{52.0} & \textbf{73.6} & \textbf{57.2} & \textbf{45.5} & \textbf{70.6} & \textbf{49.0} & \textbf{54.2} & \textbf{75.2} & \textbf{59.4} & \textbf{47.6} & \textbf{72.7} & \textbf{51.6}  \\ \hline
ViT-L\textsuperscript{$\dagger$}~\cite{li2021benchmarking}         & 337M & 45.7 & 68.9 & 49.4 & 41.5 & 65.6 & 44.6 & 48.3 & 70.4 & 52.9 & 43.4 & 67.9 & 46.6 \\
ViTDet-L\textsuperscript{$\dagger$}~\cite{li2022vitdet}      & 351M & 46.2 & 69.2 & 50.3 & 41.4 & 65.8 & 44.1 & 49.1 & 71.5 & 53.8 & 44.0 & 68.5 & 47.6 \\
ViT-Adapter-L\textsuperscript{$\dagger$}~\cite{chen2022vit-adapter} & 348M & 48.7 & 70.1 & 53.2 & 43.3 & 67.0 & 46.9 & 52.1 & 73.8 & 56.5 & 46.0 & 70.5 & 49.7 \\
\rowcolor[HTML]{ECF4FF}
ViT-CoMer-L\textsuperscript{$\dagger$}~(ours)         &  363M & 51.4 & 73.5 & 55.7 & 45.2 & 70.3 & 48.5 &  52.9  & 73.8 & 57.5  & 46.4  & 71.1  & 50.4  \\ 
\rowcolor[HTML]{DAE8FC}
ViT-CoMer-L\textsuperscript{$\ddagger$}~(ours)         &  363M & \textbf{53.4} & \textbf{75.3} & \textbf{58.9} & \textbf{46.8} & \textbf{72.0} & \textbf{50.9} &  \textbf{55.9}  & \textbf{77.3} & \textbf{61.5}  & \textbf{49.1}  & \textbf{74.5}  & \textbf{53.5}  \\ \hline
\end{tabular}
\caption{\textbf{Object detection and instance segmentation with Mask R-CNN on COCO val2017.} ``$\dagger$'' denotes the use of ImageNet-22K pre-training, ``$\ddagger$'' denotes the use of DINOv2~\cite{oquab2023dinov2}, while the default is to use ImageNet-1K pre-training.}
%\vspace{-3mm}
\label{table_detection}
\end{table*}
% 对于跨结构融合后的特征，我们采用双向交互的方式实现ViT与CNN两个分支的特征更新。具体来说，一个完整的交互过程包括ViT特征更新与CNN特征更新两部分。对于第i个stage，在stage i开始时融合两个结构特征，通过x_i' = gamma*CTI(xi,fi) 将特征注入到Transformer结构中，其中gamma是一组可学习变量，初始化为零，最小化vit在训练初期受随机初始化的cnn结构的影响。 然后在stage i 结束时提取vit的输出，重复过程，注入CNN结构。其中，stage 的数目i是根据ViT结构的深度设定的。 通过多个stage的交互，能够确保CNN与Transformer的特征在全局范围内保持紧密联系。通过跨结构的特征融合以及双向交互方式，可以充分利用不同尺度、不同层级的特征信息，为模型提供了更强的表达能力和泛化能力。从而在密集预测任务中有更好的表现。
% \subsubsection{Bidirectional Interaction.}
Regarding features fused across architectures, bidirectional interaction is employed to update features of the ViT and CNN branches. Specifically, for the $i$-th stage, at the beginning of stage $i$, the two branch features are fused, and then the fused features are injected into the ViT branch. The process can be formulated as:
\begin{equation}
    \hat{X_i} = \alpha * {O}_{2i-1} + X_i,
\end{equation} where $\hat{X_i}$ is the updated feature of the ViT branch, $\alpha$ is a learnable variable initialized to zero, it minimizes the influence of randomly initialized CNN architecture on ViT during early training. At the end of stage $i$, the process is repeated to inject features into the CNN branch, represented as:
\begin{equation}
    \hat{F_i} = {O}_{2i} + {F_i}',
\end{equation} where $\hat{F_i}$ is the updated feature of the CNN branch, the number of stages $i$ is determined based on the depth of the ViT. The cross-architecture feature fusion and bidirectional interaction enable the utilization of features from multi-scales and multi-levels, enhancing the model's expressive and generalization abilities. Simultaneously, the proposed components might be easily integrated into other advanced models and perform better in dense prediction tasks. 
% \vspace{-0.5mm}
\section{Experiment}
We select typical tasks in dense prediction: object detection, instance segmentation, and semantic segmentation and conduct extensive experiments (with different model sizes, algorithm frameworks, and configurations) on COCO~\cite{lin2014ms-coco} and ADE20K~\cite{zhou2019semantic} datasets, to verify the effectiveness of ViT-CoMer. Meanwhile, we use various pre-training of ViT, including weights pre-trained on ImageNet-1K, ImageNet-22K, and multi-modal data. ViT-CoMer achieves results that are superior to existing SOTA ViT-based methods (e.g., ViTDet~\cite{li2022vitdet}, ViT-Adapter~\cite{chen2022vit-adapter}) and comparable to vision-specific advanced methods. In addition, we perform ablation experiments on the designed modules and qualitative experiments for dense prediction tasks. These results indicate that ViT-CoMer can promote plain ViT to attain superior performance, and can be migrated as a robust backbone to various dense prediction task frameworks.
\begin{table}[t]
\centering %表格相对于caption居中
\small
\setlength{\tabcolsep}{1pt} % 调整列间距
\begin{tabular}{lcccc}
\hline
Method             & $ \mathrm{AP^{b}} $ & $ \mathrm{AP^{b}_{50}} $ & $ \mathrm{AP^{b}_{75}} $ & \#Param  \\ \hline
\multicolumn{5}{c}{Cascade Mask R-CNN 3× +MS schedule}                            \\
Swin-T~\cite{liu2021swin}             & 50.5                & 69.3                     & 54.9                     & 86M  \\
Shuffle-T~\cite{huang2021shuffle-T}          & 50.8                & 69.6                     & 55.1                     & 86M  \\
PVTv2-B2~\cite{wang2022pvtv2}           & 51.1                & 69.8                     & 55.3                     & 83M  \\
Focal-T~\cite{yang2021focal}            & 51.5                & 70.6                     & 55.9                     & 87M  \\
Swin-S~\cite{yang2021focal}            & \textbf{51.9}                & \textbf{70.7}                     & 56.3                     & 107M  \\
ViT-S~\cite{li2021benchmarking}              & 47.9                & 67.1                     & 51.7                     & 82M  \\
ViT-Adapter-S~\cite{chen2022vit-adapter}      & 51.5                & 70.1                     & 55.8                     & 86M  \\
\rowcolor[HTML]{ECF4FF}
ViT-CoMer-S (ours) &       \textbf{51.9}         &       70.6          &     \textbf{56.4}                 &  89M    \\ \hline
\multicolumn{5}{c}{ATSS 1× schedule}                                          \\
ViT-T~\cite{li2021benchmarking}              & 34.8                & 52.9                     & 36.9                       & 14M  \\
ViT-Adapter-T~\cite{chen2022vit-adapter}      & 39.3                & 57.0                     & 42.4                       & 16M  \\
\rowcolor[HTML]{ECF4FF}
ViT-CoMer-T (ours) &      \textbf{40.4}          &     \textbf{58.4}                &   \textbf{43.6}                   &   17M   \\ \hline
\multicolumn{5}{c}{GFL 1× schedule}                                           \\
ViT-T~\cite{li2021benchmarking}              & 35.7                & 53.6                     & 38.1                     & 14M  \\
ViT-Adapter-T~\cite{chen2022vit-adapter}      & 40.3                & 58.2                     & 43.4                     & 16M  \\
\rowcolor[HTML]{ECF4FF}
ViT-CoMer-T (ours) &     \textbf{40.7}     &      \textbf{58.9}       &    \textbf{43.7}     &  17M    \\ \hline
\end{tabular}
\caption{\textbf{Object detection with different frameworks on COCO val2017.} ``+MS'' means multi-scale training.}
\label{table_detection_different_framework}
\end{table}
\subsection{Object Detection and Instance Segmentation}
\indent \textbf{Settings.} We utilize the MMDetection~\cite{chen2019mmdetection} framework to implement our method and perform object detection and instance segmentation experiments on the COCO dataset. The object detection and instance segmentation frameworks encompass Mask R-CNN, Cascade Mask R-CNN, ATSS, and GFL. Referring to PVT, we conduct experiments with a training schedule of 1× (12 epochs) or 3× (36 epochs). We use a total batch size of 16, utilize the AdamW optimizer with a learning rate of $ 1\times 10^{-4} $ and a weight decay of 0.05.

\textbf{Comparisons with different backbones and frameworks.} Table~\ref{table_detection} shows the comparisons between ViT-CoMer and various scales of plain ViT, vision-specific and adapted backbones on Mask R-CNN 1× and 3× schedules. It can be seen that under similar model sizes, ViT-CoMer outperforms other backbones in the two typical dense prediction tasks of COCO object detection and instance segmentation. For instance, ViT-CoMer-S demonstrates a notable increase of +5.6\% (+4.8\%) in box mAP and +3.4\% (+3.1\%) in mask mAP compared to plain ViT-S under the 1× (3×) schedule. 
%Furthermore, in comparison to ViT-Adapter-S, which is the previous adapted-backbone-based state-of-the-art method, ViT-CoMer-S achieves +1.1\% (+0.6\%) higher box mAP and +0.6\% (+0.2\%) higher mask mAP with 1× (3×) schedule. 
ViT-CoMer-S achieves superior detection results compared to ViT-L while utilizing only $1/6$ of the parameters.
Furthermore, our approach still shows notable improvements over vision-specific and adapted backbones, such InternImage~\cite{wang2023internimage} and ViT-Adapter~\cite{chen2022vit-adapter}.

We further evaluate ViT-CoMer with different detection frameworks, the results are shown in Table~\ref{table_detection_different_framework}. It can be seen that our approach consistently outperforms other backbones across various frameworks, model sizes, and configurations.

\textbf{Results on different pre-trained weights.} We conduct experiments on Mask R-CNN (3× schedule) using different pre-trained weights, and the results are shown in Table~\ref{table_detection_different_pretrain}. Specifically, ViT-CoMer-B with multi-modal pre-training~\cite{peng2022beit}, can achieve +1.7\% $\mathrm{AP^{b}} $ and +1.7\% $\mathrm{AP^{m}} $ gain compared to ImageNet-1K~\cite{touvron2021imagenet-1k}. Furthermore, we compared more pre-training on ViT-CoMer-L, among which self-supervised pre-training~\cite{oquab2023dinov2} achieved significant results. Compared with ImageNet-22K~\cite{steiner2021imagenet-22k} pre-training, it achieves +3.0\% $\mathrm{AP^{b}} $ and +2.7\% $\mathrm{AP^{m}} $ gains. These results demonstrate that our ViT-CoMer can easily leverage diverse, open-source, large-scale pre-training to improve performance on downstream tasks.
%We conduct experiments on Mask R-CNN (3× schedule) using different pre-trained weights, the result is shown in Table~\ref{table_detection_different_pretrain}. Firstly, we initialize with the ImageNet-1K pre-training~\cite{touvron2021training}. 
%As we can see, under comparable model sizes, our ViT-CoMer-B surpasses many previous SOTA methods and achieves results comparable to those of InternImage-B, the current SOTA foundation model. Subsequently, we use the pre-trained weight~\cite{steiner2021train} obtained on a larger scale of data (ImageNet-22K), which further improves the mAP. Furthermore, we utilize a stronger multi-modal pre-training~\cite{peng2022beit} that gives a significant gain of 2.0\% $   \mathrm{AP^{b}} $ and 1.9\% $   \mathrm{AP^{m}} $ on ViT-CoMer-L. These results indicate that our approach has universality under ViT-based, open-source, large-scale, or multi-modal pre-training and can significantly outperform previous SOTA methods under a comparable number of parameters.%

% 从图6可以看出，我们的方案在不增加额外训练数据的情况下，比现有的SOTA方案获得了更好的结果。我们把原因归结于先进的预训练和本文提出的框架。

\textbf{Comparisons with state-of-the-arts.} In order to further improve the performance, we conduct experiments based on Co-DETR~\cite{codetr2022}, using ViT-CoMer as the backbone, and initializing the model with multi-modal pre-training BEiTv2.
%We conducted experiments based on the Co-DETR~\cite{codetr2022}framework, and initializing the model with multi-modal pre-training Beitv2~\cite{peng2022beit}.and utilized ViT-CoMer as the backbone, following it's experimental configuration, and initialized the model with multi-modal pre-trained BEiTv2~\cite{peng2022beit}.%
%We replaced the backbone of ViT-CoMer with a Co-DETR~\cite{codetr2022} using BEiTv2~\cite{peng2022beit} multimodal pre-trained. 
%We combine our ViT-CoMer and Co-DETR~\cite{codetr2022} frameworks with multi-modal pre-training BEiTv2~\cite{peng2022beit}.
As shown in Table~\ref{det_sota}, our approach outperforms the existing SOTA algorithms without extra training data on COCO val2017, which strongly demonstrates the effectiveness of ViT-CoMer.
%The implementation details are in the supplementary material.
%Without extra training data on COCO val2017, our method outperforms the existing SOTA algorithms, as Table ~\ref{det_sota} illustrates. 
%The success of our approach may be partially attributed to advanced pre-training and the ViT-CoMer framework.
%Utilizing readily available pre-trained ViTs models allows us to enhance our competitive edge.
\begin{table}[t]
\centering %表格相对于caption居中
\small
\setlength{\tabcolsep}{1.5pt} % 调整列间距
\begin{tabular}{c | l | cc}
\hline
Pre-training                     & \multicolumn{1}{c|}{Method}             & $   \mathrm{AP^{b}} $ & $   \mathrm{AP^{m}} $ \\
\hline
IN-1K  
                              %& Swin-B~\cite{liu2021swin}             & 48.6                  & 43.3                  \\
                              %& ViT-B~\cite{li2021benchmarking}              & 45.8                  & 41.3                  \\
                              %& InternImage-B~\cite{wang2023internimage} & \textbf{50.3} & \textbf{44.8} \\
                              %& ViT-Adapter-B~\cite{chen2022vit-adapter}      & 49.6                  & 43.6                  \\
                              & ViT-CoMer-B &   50.2                 &    44.0          \\
\hline
IN-22K %& ViT-L~\cite{li2021benchmarking}              & 48.3                  & 43.4                  \\
                              %& ViTDet-L~\cite{li2022vitdet}           & 49.1                  & 44.0                  \\
                              %& ViT-Adapter-L~\cite{chen2022vit-adapter}      & 52.1                  & 46.0                  \\
                              & ViT-CoMer-L & 52.9                  & 46.4                  \\
\hline
\multirow{2}*{MM}  %& ViT-Adapter-B~\cite{chen2022vit-adapter} & 51.2                  & 45.3                  \\
                              & ViT-CoMer-B &   51.9                    &       45.7        \\
                              & ViT-CoMer-L &    54.9             &    48.3      \\
\hline
SSL & ViT-CoMer-L & 55.9 & 49.1 \\
\hline
\end{tabular}
\caption{\textbf{Comparisons of different pre-training for object detection and instance segmentation tasks with Mask R-CNN on COCO val2017.} IN-1K, IN-22K, MM and SSL respectively represent the use of ImageNet-1K~\cite{touvron2021imagenet-1k}, ImageNet-22K~\cite{steiner2021imagenet-22k}, multi-modal, self-supervised learning pre-training.}
% \vspace{-3mm}
\label{table_detection_different_pretrain}
\end{table}
\begin{table}[]
\centering %表格相对于caption居中
\footnotesize
\setlength{\tabcolsep}{1.5pt} % 调整列间距
\begin{tabular}{lllcc}
\hline
Method               & Backbone      & Pre-training  & \#P & $   \mathrm{AP^{b}} $ \\
\hline
DINO-D-DETR~\cite{zhang2022dino-d-detr} & Swin-L~\cite{liu2021swin}        & IN-22K        & 284M    & 58.5      \\
HTC++~\cite{liu2021swin}           & ViT-Adapter-L~\cite{chen2022vit-adapter} & BEiTv2~\cite{peng2022beit}      & 401M    & 60.5      \\
HTC++~\cite{liu2021swin}           & ViT-Adapter-L~\cite{chen2022vit-adapter} & BEiTv2+O365 & 401M    & 62.6      \\
CMask R-CNN~\cite{cai2019cascade-rcnn}          & ViTDet-H~\cite{li2022vitdet}      & IN-1K         & 692M    & 61.3      \\
Co-DETR~\cite{codetr2022}              & Swin-L~\cite{liu2021swin}        & IN-22K        & 218M    & 60.7      \\
\rowcolor[HTML]{ECF4FF}
Co-DETR~\cite{codetr2022}               & ViT-CoMer-L~(ours)   & BEiTv2~\cite{peng2022beit}        & 363M    & 62.1      \\
\rowcolor[HTML]{DAE8FC}
Co-DETR~\cite{codetr2022}             & ViT-CoMer-L~(ours)   & BEiTv2\textsuperscript{$\ast$}~\cite{chen2022vit-adapter} & 363M    & 64.3   \\ \hline  
\end{tabular}
\caption{\textbf{Comparisons with previous SOTA on COCO  val 2017.} O365 indicates the Objects365 dataset is used during training. \textsuperscript{$\ast$} indicates a variant version of BEiTv2 used in ViT-Adapter.}
% \vspace{-3mm}
\label{det_sota}
\end{table}
\label{sec:different_pretrained_weights}
\begin{table}[t]
\centering %表格相对于caption居中
\small
\setlength{\tabcolsep}{4pt} % 调整列间距
\begin{tabular}{l | c  ccc}
\hline
\multirow{2}{*}{Method} & \multicolumn{3}{c}{UperNet 160k} \\
                                           & \#Param     & mIoU     & +MS      \\
\hline
PVT-T~\cite{wang2022pvtv2}                     & 43.2M       & 38.5     & 39.0    \\
ViT-T~\cite{li2021benchmarking}                 & 34.1M       & 41.7     & 42.6    \\
ViT-Adapter-T~\cite{chen2022vit-adapter}            & 36.1M       & 42.6     & 43.6    \\
\rowcolor[HTML]{ECF4FF}
ViT-CoMer-T~(ours)              &     38.7M     &   \textbf{43.0}    &    \textbf{44.3}  \\
\hline
PVT-S~\cite{wang2022pvtv2}                   & 54.5M       & 43.7     & 44.0    \\
Swin-T~\cite{liu2021swin}                             & 59.9M       & 44.5     & 45.8    \\
Twins-SVT-S~\cite{chu2021twins}                      & 54.4M       & 46.2     & 47.1    \\
ViT-S~\cite{li2021benchmarking}                            & 53.6M       & 44.6     & 45.7    \\
ViT-Adapter-S~\cite{chen2022vit-adapter}                       & 57.6M       & 46.2     & 47.1    \\
\rowcolor[HTML]{ECF4FF}
ViT-CoMer-S~(ours)                   &    61.4M       &    \textbf{46.5}   &   \textbf{47.7}   \\
\hline
Swin-B~\cite{liu2021swin}                              & 121.0M      & 48.1     & 49.7    \\
Twins-SVT-L~\cite{chu2021twins}                           & 133.0M      & \textbf{48.8}     & \textbf{50.2}    \\
ViT-B~\cite{li2021benchmarking}                             & 127.3M      & 46.1     & 47.1    \\
ViT-Adapter-B~\cite{chen2022vit-adapter}                     & 133.9M      & \textbf{48.8}     & 49.7    \\
\rowcolor[HTML]{ECF4FF}
ViT-CoMer-B~(ours)                    &      144.7M       &  \textbf{48.8}     &  49.4     \\
\hline
Swin-L\textsuperscript{$\dagger$}~\cite{liu2021swin}                            & 234.0M      & 52.1     & 53.5    \\
ViT-Adapter-L\textsuperscript{$\dagger$}~\cite{chen2022vit-adapter}                      & 363.8M      & 53.4     & 54.4    \\
\rowcolor[HTML]{ECF4FF}
ViT-CoMer-L~(ours)\textsuperscript{$\dagger$}                   & 383.4M      & \textbf{54.3}     & \textbf{55.6}    \\
\hline
\end{tabular}
\caption{\textbf{Semantic segmentation results on the ADE20K val. ``+MS'' means multi-scale testing.} ``$\dagger$'' denotes the use of ImageNet-22K pre-trained weight, while the default is to use ImageNet-1K pre-training.}
\label{table_segmentation}
\end{table}

\begin{table}[]
\centering %表格相对于caption居中
\small
\setlength{\tabcolsep}{1.5pt} % 调整列间距
\begin{tabular}{c|l|cc}
\hline
Pre-training & \multicolumn{1}{c|}{Method}        & mIoU & +MS   \\ \hline
          %& Twins-SVT-S~\cite{chu2021twins}   & 46.2 & 47.1 \\
          %& ViT-S~\cite{li2021benchmarking}         & 44.6 & 45.7 \\
          %& ViT-Adapter-S~\cite{chen2022vit-adapter} & 46.2 & 47.1 \\
%IN-1K  & ViT-CoMer-S & 46.5 & 47.7 \\ \hline
\multirow{3}*{IN-22K}          & Swin-L~\cite{liu2021swin}        & 52.1 & 53.5 \\
          & ViT-Adapter-L~\cite{chen2022vit-adapter} & 53.4 & 54.4 \\ 
 & \cellcolor[HTML]{ECF4FF}ViT-CoMer-L~(ours) & \cellcolor[HTML]{ECF4FF}\textbf{54.3} & \cellcolor[HTML]{ECF4FF}\textbf{55.6}     \\ \hline
\multirow{2}*{MM}          & ViT-Adapter-L~\cite{chen2022vit-adapter} & 55.0 & 55.4 \\
  & \cellcolor[HTML]{ECF4FF}ViT-CoMer-L~(ours) & \cellcolor[HTML]{ECF4FF}\textbf{56.3} & \cellcolor[HTML]{ECF4FF}\textbf{56.8}     \\ \hline
\end{tabular}
\caption{\textbf{Comparisons of different pre-trained weights for semantic segmentation with UperNet on ADE20K val.} IN-22K and MM respectively represent the use of ImageNet-22K and multi-modal pre-trained weights.}
\label{seg_dif_pretrain}
% \vspace{-3mm}
\end{table}

\subsection{Semantic Segmentation}
\textbf{Settings.} Our semantic segmentation experiments are based on the ADE20K dataset with MMSegmentation~\cite{contributors2020mmsegmentation}. We select UperNet~\cite{xiao2018unified} as the basic framework. The training configuration remains consistent with Swin~\cite{liu2021swin}, encompassing training for 160,000 iterations. The batch size is set to 16, and the AdamW optimizer is used. The learning rate and weight decay parameters are tuned to $ 2\times 10^{-5} $ and 0.05, respectively.

\textbf{Comparisons with different backbones.} Table~\ref{table_segmentation} presents the comparisons of both single-scale and multi-scale mIoU between ViT-CoMer and various backbones, including plain ViT, vision-specific backbones, and adapted backbones in semantic segmentation tasks. It shows that, under comparable model sizes, our method surpasses the ViT and many vision-specific backbones. For instance, our ViT-CoMer-S achieves 47.7\% MS mIoU, outperforming many strong counterparts such as Swin-T (+1.9\%) and ViT-Adapter-S (+0.6\%). Similarly, ViT-CoMer-L reports a competitive performance of 55.6\% MS mIoU, which is 2.1\% higher than Swin-L and 1.2\% higher than ViT-Adapter-L. These equitable comparisons demonstrate the effectiveness and universality of our ViT-CoMer in the semantic segmentation task.

\textbf{Comparisons with different pre-trained weights.} Table~\ref{seg_dif_pretrain} is the result of using different pre-trained weights on UperNet. 
%We first initialize models with the DeiT~\cite{touvron2021training} released ImageNet-1K weights (without distillation). 
%It shows that, under comparable model sizes, our method surpasses the plain ViT, many vision-specific transformers, and previous SOTA adapted-backbone-based ViT-Adapter. 
When using the ImageNet-22K pre-trained weights~\cite{steiner2021train}, our ViT-CoMer-L attains 55.6\% MS mIoU, exceeding ViT-Adapter-L by 1.2\% mIoU. Then, we initialize ViT-CoMer-L with the multi-modal pre-training~\cite{zhu2022uni-perceiver}, which benefits our model with impressive gains of 2.0\% mIoU, exceeding ViT-Adapter-L by 1.4\%. These significant and consistent improvements suggest that our method can effectively improve plain ViT and fully utilize various open-source ViT-based pre-trained weights, enabling the model to perform better in semantic segmentation.

\textbf{Comparisons with state-of-the-arts.} To enhance the performance even more, we conduct experiments based on Mask2Former~\cite{cheng2021mask2former} using ViT-CoMer as the backbone, and initializing the model with multi-modal pre-training BEiTv2. As shown in Table~\ref{seg_sota}, our method achieves comparable performance to SOTA methods on ADE20K with fewer parameters.
\begin{table}[]
\centering %表格相对于caption居中
\footnotesize
\setlength{\tabcolsep}{1.5pt} % 调整列间距
\begin{tabular}{llccccc}
\hline
Method      & Backbone      & Pre-train & \#P & mIoU & +MS  \\
\hline
Mask DINO~\cite{li2022maskdino}   & Swin-L~\cite{liu2021swin}        & IN-22K       & 223M             & 59.5 & 60.8 \\
Mask2Former~\cite{cheng2021mask2former} & ViT-Adapter-G~\cite{chen2022vit-adapter} & BEiTv3~\cite{beit3}       & 1.9B           & 62.0 & 62.8 \\
Mask2Former~\cite{cheng2021mask2former} & ViT-Adapter-G~\cite{chen2022vit-adapter} & EVA~\cite{EVA}          & 1.0B           & 61.5 & 62.3 \\
Mask2Former~\cite{cheng2021mask2former} & RevCol-H~\cite{cai2022revcol}      & -            & 2.4B          & 60.4 & 61.0 \\
Mask2Former~\cite{cheng2021mask2former} & ViT-Adapter-L~\cite{chen2022vit-adapter} & BEiTv2~\cite{peng2022beit}       & 571M           & 61.2 & 61.5 \\
\rowcolor[HTML]{ECF4FF}
Mask2Former~\cite{cheng2021mask2former} & ViT-CoMer-L~(ours)   & BEiTv2~\cite{peng2022beit}       & 604M          & 61.7 & 62.1 \\
\hline
\end{tabular}
\caption{\textbf{Comparisons with previous SOTA on ADE20K dataset for semantic segmentation.}}
\label{seg_sota}
\end{table}
\begin{table}[]
\centering %表格相对于caption居中
%\small
\setlength{\tabcolsep}{2pt} % 调整列间距
\begin{tabular}{l | ccc | cc}
\hline
                   & \multicolumn{3}{c|}{Components} &      &      \\
\multirow{-2}{*}{Method} & MRFP & CTI (to V) & CTI (to C) & \multirow{-2}{*}{$   \mathrm{AP^{b}} $} & \multirow{-2}{*}{$   \mathrm{AP^{m}} $} \\
\hline
\multirow{4}*{ViT-S}        &    ×     &    ×     &     ×    & 40.2 & 37.1 \\
          & \checkmark        &     ×    &     ×    & 41.5 & 38.2 \\
          & \checkmark        & \checkmark        &    ×     & 43.3 & 39.3 \\ 
 & \cellcolor[HTML]{ECF4FF}\checkmark        & \cellcolor[HTML]{ECF4FF}\checkmark        & \cellcolor[HTML]{ECF4FF}\checkmark        & \cellcolor[HTML]{ECF4FF}\textbf{45.8} & \cellcolor[HTML]{ECF4FF}\textbf{40.5}   \\
\hline
\end{tabular}
\caption{\textbf{Ablation studies of key components.} Our proposed components collectively bring 5.6 $   \mathrm{AP^{b}} $ and 3.4 $   \mathrm{AP^{m}} $ gains. CTI (to V) indicates that the fused features are injected into the ViT branch, whereas CTI (to C) means that the fused features are injected into the CNN branch.}
\label{table_ablation_components}
\end{table}
\subsection{Ablation Study}
\textbf{Settings.} We conduct ablation experiments on the ViT-CoMer-S, using Mask R-CNN (1× schedule) for object detection and instance segmentation tasks. The total batch size used during the training process is 16, the optimizer employed is AdamW, and the learning rate and weight decay parameters are set to $ 1\times 10^{-4} $ and 0.05, respectively.

\textbf{Ablation for components.} We gradually add the proposed submodules to the ViT-S, ultimately evolving the model into the ViT-CoMer. The results of the ablation experiment are shown in Table~\ref{table_ablation_components}. When MRFP is used to provide multi-scale and multi-receptive-field features of CNN to plain ViT (features are directly added), it results in improvements of 1.3\% $   \mathrm{AP^{b}} $ and 1.1\% $   \mathrm{AP^{m}} $. Furthermore, we replace the ``directly added'' 
operation with CTI proposed in this work. When only CTI (to V) is used, the model improves by 1.8\% $   \mathrm{AP^{b}} $ and 1.1\% $   \mathrm{AP^{m}} $; when CTI (to V) and CTI (to C) are used simultaneously, the performance further significantly improves by 2.5\% $   \mathrm{AP^{b}} $ and 1.2\% $   \mathrm{AP^{m}} $. Overall, compared to plain ViT, our ViT-CoMer achieved significant improvements of 5.6\% $   \mathrm{AP^{b}} $ and 3.4\% $   \mathrm{AP^{m}} $. The experimental results demonstrate that our proposed MRFP and CTI modules can significantly enhance the ability of plain ViT, making it well adapted to dense prediction tasks.

\textbf{Number of bidirectional fusion interaction.} In Table~\ref{table_ablation_n}, we analyze the impact of the number of bidirectional fusion interaction modules. We observe that as N increases, the model accuracy reaches a plateau, and introducing more interactions does not consistently enhance performance. Therefore, we set N to 4 by default.

\textbf{Different kernel size in MRFP.} Table~\ref{table_ablation_k} illustrates the influence of varying kernel sizes on the MRFP. The results show that the number of parameters increases as the kernel size increases. Simultaneously, we observe $   \mathrm{AP^{b}} $ and $   \mathrm{AP^{m}} $ peaks when using kernel sizes 3 and 5, therefore we adopt these as the default settings.
\begin{table}[]
\centering %表格相对于caption居中
%\small
\begin{tabular}{l | ccc}
\hline
N & $   \mathrm{AP^{b}} $  & $   \mathrm{AP^{m}} $  & \#Param \\
\hline
0 & 40.2 & 37.1 & 43.8M   \\
2 & 45.1     &   40.1   &   48.2M      \\
\rowcolor[HTML]{ECF4FF} 
4 & \textbf{45.8} & \textbf{40.5} & 50.3M   \\
6 &   45.6   &  \textbf{40.5}    &  52.5M       \\
\hline
\end{tabular}
\caption{\textbf{Ablation of the number of bidirectional fusion interaction modules.} The model performs best when $N$=4.}
\label{table_ablation_n}
% \vspace{-3mm}
\end{table}
\begin{table}[]
\centering %表格相对于caption居中
%\small
\begin{tabular}{c | ccc}
\hline
k & $   \mathrm{AP^{b}} $  & $   \mathrm{AP^{m}} $  & \#Param \\
\hline
3 & 45.7 & 40.4 & 50.29M   \\
\rowcolor[HTML]{ECF4FF} 
3, 5 & \textbf{45.8}  &   \textbf{40.5}  &   50.31M      \\
3, 5, 7 & 45.5 & 40.2 & 50.33M   \\
3, 5, 7, 9 &   45.4   &  40.0    &  50.36M       \\
\hline
\end{tabular}
\caption{\textbf{Ablation of the setting of kernel size in MRFP .} The model performs best when $k$=3 and 5.}
\label{table_ablation_k}
\vspace{-3mm}
\end{table}
\subsection{Scalability}
Our method also can be employed with hierarchical vision transformers such as Swin. We apply our approach to Swin-T with Mask R-CNN (1x schedule) for object detection. As illustrated in Table~\ref{table_swin}, our method improves the performance of Swin-T by +2.1\% box AP and +1.2\% mask AP. Since the Swin architecture already introduces inductive biases, the improvements are somewhat lower compared to a plain ViT. Nonetheless, these results still substantiate the scalability of our approach.
\subsection{Qualitative Results}
\begin{figure}[t]
  \centering
  \includegraphics[width=1.0\linewidth]{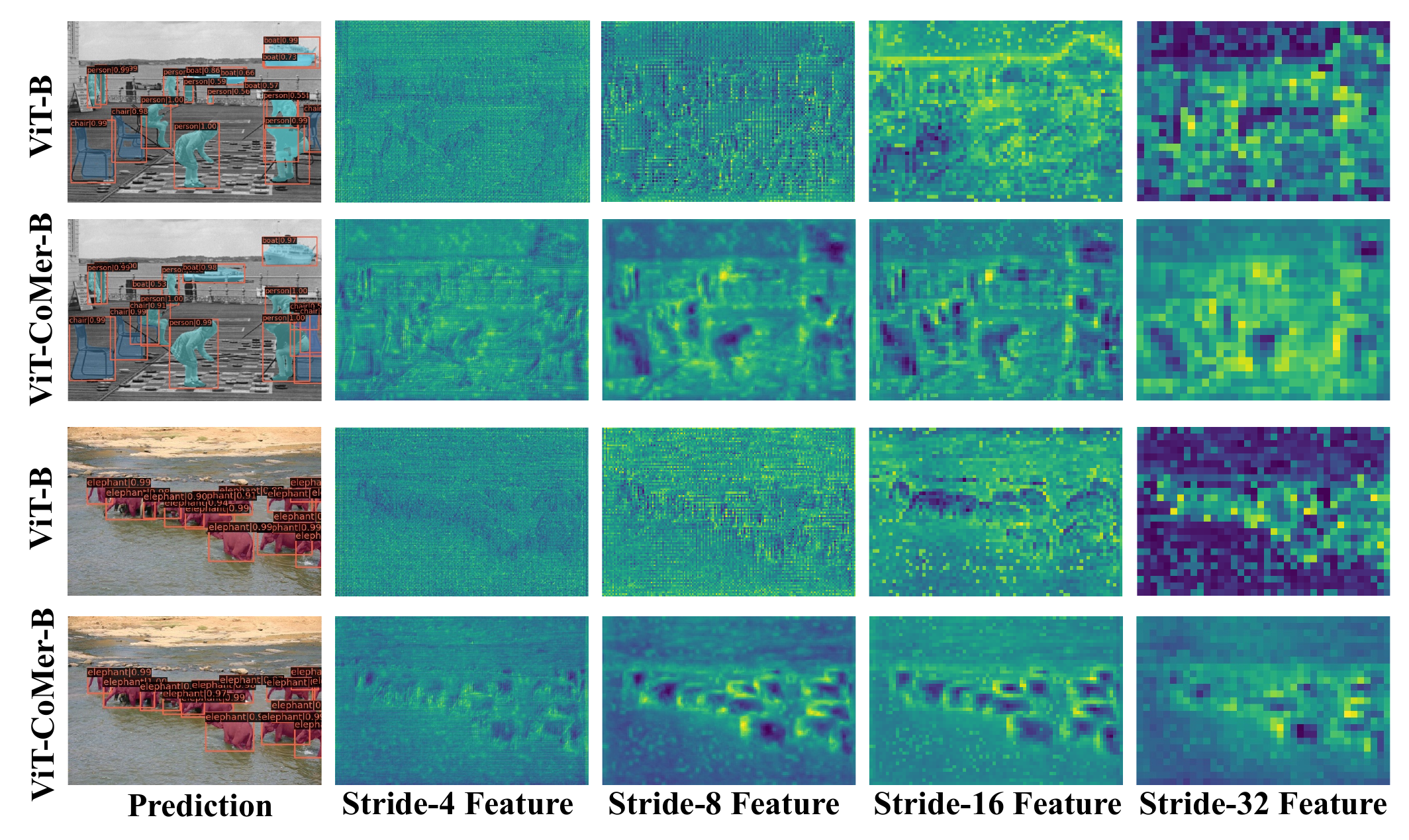}
   \caption{\textbf{Visualization of feature maps for object detection and instance segmentation.} Prediction results and feature maps with different resolutions are generated from ViT-B and ViT-CoMer-B.}
   \label{figure_visualization}
% \vspace{-2mm}
\end{figure}

According to iFormer~\cite{si2022iformer}, plain ViT tends to capture global and low-frequency features in images due to self-attention operations, while CNN tends to capture local and high-frequency features in the image due to convolution operations. However, in dense prediction tasks, various objects will appear in the image with different sizes and densities, which requires the model to have the ability to simultaneously extract and capture local and global, high-frequency and low-frequency features. We qualitatively evaluate the difference between plain ViT and our proposed ViT-CoMer by visualizing feature maps on different layers (downsampling $1/4$, $1/8$, $1/16$, and $1/32$) for instance segmentation and object detection tasks. The qualitative visualization results are shown in Figure~\ref{figure_visualization}. It can be seen that compared to the plain ViT, our ViT-CoMer yields more fine-grained multi-scale features, thereby enhancing the model's object localization capability.
\begin{table}[]
\centering %表格相对于caption居中
%\small
\begin{tabular}{c | ccc}
\hline
Method & $   \mathrm{AP^{b}} $  & $   \mathrm{AP^{m}} $ & \#Param \\
\hline
Swin-T & 42.7 & 39.3 & 48M   \\
\rowcolor[HTML]{ECF4FF} 
Swin-CoMer-T~(ours)  &   \textbf{44.8}  &   \textbf{40.5} & 54M   \\
\hline
\end{tabular}
\caption{\textbf{Scalability of the Swin Transformer.}}
% \vspace{-5mm}
\label{table_swin}
\end{table}
\section{Conclusion}
% 在这篇工作中，我们提出了ViT-CoMer, 一种plain, non-hieralchicle, 特征增强的vit 框架。该框架充分利用了卷积和transformer的优势。在不改变vit框架的情况下，增加了多尺度卷积特征交互模块重构细粒度层级语义特征。我们在在目标检测、实例分割和语义分割等密集任务验证了vit-comer。大量实验证明，我们的方案可以获的比同类型框架更好的预测结果，同时获得和vision-specific框架方案可比甚至更优的预测效果。
In this work, we propose ViT-CoMer, a plain, non-hierarchical, and feature-enhanced ViT backbone that effectively leverages the strengths of both CNN and Transformer. Without altering the ViT architecture, we integrate a multi-scale convolutional feature interaction module to reconstruct fine-grained hierarchical semantic features. We validate ViT-CoMer on dense prediction tasks including object detection, instance segmentation, and semantic segmentation. Extensive experiments demonstrate that our approach can achieve superior performance compared to both plain and adapted backbones. Additionally, our approach can easily obtain advanced ViT pre-trained weights and attain comparable, even surpassing performance compared to state-of-the-art backbones.
{
    \small
    \bibliographystyle{ieeenat_fullname}
    \bibliography{paper}
}


\end{document}